\documentclass[runningheads]{llncs}
\usepackage[T1]{fontenc}
\usepackage{graphicx}
\usepackage{booktabs}
\usepackage{multirow}
\usepackage{array}
\usepackage{subcaption}
\usepackage{amsmath}
\usepackage{amssymb}
\usepackage{url}
\usepackage{microtype}
\usepackage[unicode,
    colorlinks=true,
    linkcolor=blue,
    urlcolor=blue,
    citecolor=blue,
    bookmarks=true,
    bookmarksnumbered=true]{hyperref}

\begin{document}

\title{Class-Dependent Perturbation Effects in Evaluating Time Series Attributions}
\titlerunning{Class-Dependent Perturbation Effects}
\authorrunning{Baer et al.}

\author{Gregor Baer\inst{1, 2}\orcidID{0009-0002-9918-1376} \and
Isel Grau\inst{1, 2}\orcidID{0000-0002-8035-2887} \and
Chao Zhang\inst{2, 3}\orcidID{0000-0001-9811-1881} \and
Pieter Van Gorp \inst{1,2}\orcidID{0000-0001-5197-3986}}

\institute{Information Systems, Eindhoven University of Technology, Eindhoven, The Netherlands \and Eindhoven Artificial Intelligence Systems Institute, Eindhoven University of Technology, Eindhoven, The Netherlands \and Human-Technology Interaction, Eindhoven University of Technology, Eindhoven, The Netherlands
}

\maketitle

\begin{abstract}
As machine learning models become increasingly prevalent in time series applications, Explainable Artificial Intelligence (XAI) methods are essential for understanding their predictions. Within XAI, feature attribution methods aim to identify which input features contribute the most to a model's prediction, with their evaluation typically relying on perturbation-based metrics. Through systematic empirical analysis across multiple datasets, model architectures, and perturbation strategies, we reveal previously overlooked class-dependent effects in these metrics: they show varying effectiveness across classes, achieving strong results for some while remaining less sensitive to others. In particular, we find that the most effective perturbation strategies often demonstrate the most pronounced class differences.  Our analysis suggests that these effects arise from the learned biases of classifiers, indicating that perturbation-based evaluation may reflect specific model behaviors rather than intrinsic attribution quality. We propose an evaluation framework with a class-aware penalty term to help assess and account for these effects in evaluating feature attributions, offering particular value for class-imbalanced datasets. Although our analysis focuses on time series classification, these class-dependent effects likely extend to other structured data domains where perturbation-based evaluation is common.\footnote{Code and results are available at~\url{https://github.com/gregorbaer/class-perturbation-effects}.}
.
\end{abstract}

\section{Introduction}

Explainable Artificial Intelligence (XAI) has emerged as a critical paradigm for understanding complex machine learning models, particularly in domains where trust and explainability are essential, such as finance or healthcare. Within XAI, feature attribution methods quantify how input features contribute to model predictions, with their often model-agnostic nature enabling application across different architectures and data types. These methods are increasingly being applied to structured data domains, such as time series, where temporal dependencies pose unique challenges. In such contexts, ensuring reliable evaluation of attribution quality becomes crucial~\cite{theissler.etal_2022_explainable}.

The evaluation of feature attribution methods faces a fundamental methodological challenge: the absence of a ground truth for explanations. Although human-centered evaluation offers a direct assessment of the utility of explanations~\cite{rong.etal_2024_humancentered}, it suffers from scalability limitations and potential domain-specific biases. Consequently, functional evaluation approaches have emerged as primary validation frameworks, with the aim of computationally verifying whether attribution methods satisfy certain desirable properties~\cite{doshi-velez.kim_2018_considerations,nauta.etal_2023_anecdotal}. Perturbation analysis represents one such framework that evaluates attribution correctness by measuring how modifying features impacts model predictions. This approach rests on a key assumption: perturbing important features should yield proportional changes in model output. 

Although perturbation analysis has gained traction for evaluating attribution methods in structured data domains like time series, previous work has mainly focused on aggregate performance metrics. Studies note that perturbation effectiveness can vary substantially with data characteristics, leading to recommendations to evaluate multiple ways of perturbing features~\cite{schlegel.keim_2023_deep,serramazza.etal_2024_improving}. However, how this effectiveness varies with specific data characteristics remains largely unexplored. A closer examination of reported results reveals an intriguing pattern: substantial portions of datasets can remain unaffected by perturbation when using a single strategy uniformly across all instances~\cite{schlegel.keim_2023_deep,simic.etal_2022_perturbation}. This observation suggests underlying methodological challenges that have not yet been systematically investigated.

Our analysis of these empirical patterns points to an important methodological limitation: the effectiveness of perturbation-based evaluation can vary substantially across different predicted classes, which we refer to as \emph{class-dependent perturbation effects}. These effects manifest when perturbation strategies effectively validate feature attributions for some classes while showing limited or no sensitivity for others. We hypothesize that such behavior emerges from classifier biases, where models learn to associate certain perturbation values with specific classes, potentially compromising the reliability of current evaluation practices.

Our research examines how class-dependent effects influence perturbation-based evaluation of attributions. Through extensive empirical analysis, we show that these effects appear more pronounced with perturbation strategies that show strong aggregate performance, and persist across different perturbation strategies, model architectures, and attribution methods. This asymmetry in perturbation effectiveness has important implications: data set imbalance may influence evaluation results, and evaluation metrics might reflect specific model behaviors rather than attribution quality. Although our evidence stems from time series classification, similar considerations may extend to other structured data domains such as computer vision.

This paper makes several contributions to existing XAI evaluation methodologies. First, we identify and characterize class-dependent effects in the evaluation of feature attributions with perturbation, supported by comprehensive empirical evidence across four commonly used benchmark datasets. Second, we introduce a penalty term that can be applied to any aggregate XAI evaluation metric to investigate the extent of class-dependent effects. Third, we provide recommendations for evaluation protocols that consider these effects, including how to assess whether attribution methods are affected by class-specific perturbation behaviors.

The remainder of this paper is organized as follows. Section~\ref{h:related-work} discusses related work on perturbation analysis for time series classification.
Section~\ref{h:method} introduces the notation and a formal definition of perturbation analysis, as well as the metrics used to measure explanation correctness and class differences.
Section~\ref{h:experiment-set-up} presents our experimental setup to investigate class-dependent perturbation effects. 
Section~\ref{h:results-discussion} analyzes our results and discusses their implications for XAI evaluation. 
Finally, Section~\ref{h:conclusion} concludes with recommendations for future research directions.

\section{Related Work}\label{h:related-work}

The evaluation of explanations remains a critical challenge in XAI research. To address this, functional evaluation techniques have emerged as key computational methods for assessing the quality of explanations without human intervention~\cite{doshi-velez.kim_2018_considerations,nauta.etal_2023_anecdotal}. 

We focus on perturbation analysis as a computational method to measure the correctness and compactness of explanations, properties identified as key quality criteria for XAI methods~\cite{nauta.etal_2023_anecdotal}. This approach was first introduced by Samek et al.~\cite{samek.etal_2017_evaluating} to evaluate feature attribution methods in the image domain. It involves sequentially perturbing pixels in order of the most relevant features first by replacing them with noninformative values and observing the impact on model predictions. This process generates a perturbation curve that tracks these prediction changes, allowing the calculation of metrics such as the area under or over the curve to jointly measure the correctness and compactness of explanations.

In this context, correctness refers to whether an explanation faithfully identifies features that truly influence the model's prediction. Compactness captures how concisely the explanation represents the model's behavior. A compact explanation would show a quick degradation of predictions when perturbing just a few highly-ranked features, indicating that the model relies on a small subset of input features. The fundamental assumption underlying this approach is that perturbing important features should degrade model predictions proportionally to their attributed importance, while perturbing irrelevant features should have minimal effects on the model output.

\begin{figure}[!htbp]
\includegraphics[width=\textwidth]{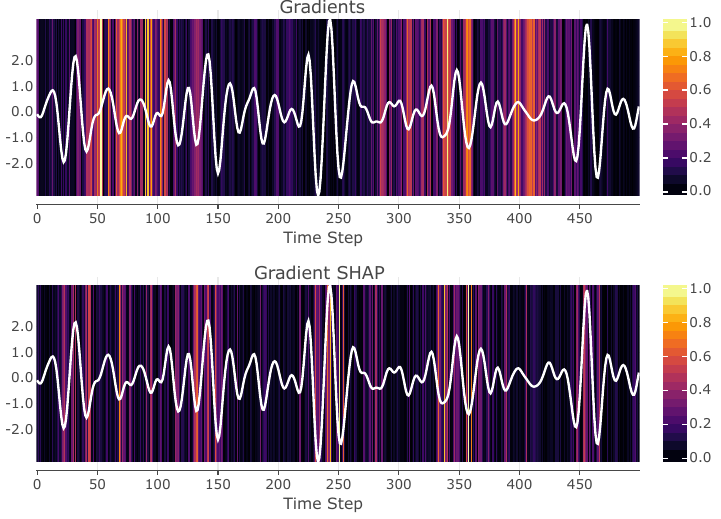}
\caption{Output from Gradients~\cite{simonyan.etal_2014_deep} and Gradient SHAP~\cite{lundberg.lee_2017_unified} attribution methods for InceptionTime~\cite{ismailfawaz.etal_2020_inceptiontime} classifier on one FordB dataset sample. The white line represents the input time series, while the heatmap indicates feature importance for the predicted class over time, with lighter colors denoting higher importance. Attributions were normalized to [0,1].}\label{fig:example_attributions}
\end{figure}

Within time series classification, there are various explanation methods, categorized into approaches based on time points, subsequences, instances, and others~\cite{theissler.etal_2022_explainable}. Our work focuses on feature attribution at the level of time points, examining how each point within a time series contributes to model predictions. As illustrated in Figure~\ref{fig:example_attributions}, these explanations identify the most influential parts of a time series for a prediction, visualized as a heatmap where darker regions indicate minimal contribution and lighter regions indicate stronger contribution to the prediction. The figure also demonstrates how different attribution methods can yield varying explanations for the same instance, highlighting the need for robust evaluation methods that answer the question of which explanation is correct.

\begin{figure}[!htbp]
\includegraphics[width=\textwidth]{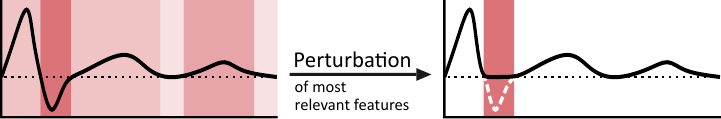}
\caption{Schematic illustration of time series perturbation based on feature attributions. Reddish rectangles represent feature attribution importance (darker red indicates higher importance). This example shows one perturbation approach: replacing values in the most important region with a constant value (e.g. zero).}\label{fig:perturbation-illustration}
\end{figure}

Figure~\ref{fig:perturbation-illustration} demonstrates how feature attributions can guide the perturbation process, where important time points or segments of a time series (indicated by darker red) are replaced with non-informative values like zero. This approach forms the basis of perturbation-based evaluation methods discussed below.
Schlegel et al.~\cite{schlegel.etal_2019_rigorous} were the first to apply perturbation analysis to time series classification, evaluating the quality of attributions by measuring the average change in accuracy over the perturbed samples with four perturbation strategies, including zero and mean value replacement.
Mercier et al.~\cite{mercier.etal_2022_time} expanded on this framework by evaluating attributions with additional metrics from image explanations, such as sensitivity and infidelity, revealing that no single attribution method consistently outperforms others in all aspects of evaluation.

Another methodological advance came from Šimić et al.~\cite{simic.etal_2022_perturbation}, who introduced a metric that compares perturbations of features ordered by their attributed importance. Their approach addressed a key limitation: perturbation strategies can affect predictions regardless of the relevance and location of the feature. By measuring the difference between most and least relevant feature perturbations, they provided a more robust assessment of attribution quality. Their metric builds upon the degradation score introduced for image explanations~\cite{schulz.etal_2020_restricting}, adapting it with cubic weighting to emphasize the early divergence between perturbation orders.

Further developments focused on understanding the effectiveness of perturbations. 
Schlegel et al.~\cite{schlegel.keim_2023_deep} introduced a novel visualization method to qualitatively assess the effectiveness of perturbations by showing class distribution histograms and distances between the original and perturbed time series, among others. They also benchmarked the effectiveness of 16 perturbation strategies by recording the number of flipped class labels.
Building on this, Schlegel et al.~\cite{schlegel.keim_2023_introducing} introduced the AttributionStabilityIndicator, which incorporates the correlation between original and perturbed time series to ensure minimal data perturbations while maintaining a significant prediction impact.

Recent work explored additional methodological refinements. Turbé et al.~\cite{turbe.etal_2023_evaluation} incorporated perturbations into model training to mitigate distribution shifts, while Nguyen et al.~\cite{nguyen.etal_2024_robust} developed a framework to recommend optimal explanation methods based on aggregate accuracy loss across perturbed samples.
Furthermore, Serramazza et al.~\cite{serramazza.etal_2024_improving} evaluated and extended InterpretTime~\cite{turbe.etal_2023_evaluation} on various multivariate time series classification tasks by averaging different perturbation strategies and applying said strategies in chunks.

However, an important question has remained unexplored: how do the characteristics of different classes affect perturbation-based evaluation methods? Perturbation strategies may be influenced by classifier biases. For example, if a classifier learned to associate certain perturbation values (such as zero) with specific classes, the effectiveness of perturbation-based evaluation could vary between different predicted classes. 
This phenomenon occurs when perturbation values inadvertently match features the model has associated with a specific class. As a result, substituting ``important'' features with these values might paradoxically reinforce rather than disrupt the prediction, regardless of attribution correctness.

This consideration may help explain some findings in the literature. For example, Schlegel et al.~\cite{schlegel.keim_2023_deep} evaluated 16 different perturbation approaches and found that for most datasets, only up to 60\% of the samples changed their predicted label under perturbation, regardless of the strategy employed. These results suggest that the effectiveness of perturbations might be influenced by factors beyond the perturbation strategy itself. Our work investigates whether class-dependent effects could explain these observed patterns, examining how the relationship between perturbation strategies and learned class representations might affect evaluation outcomes.

\section{Class-Adjusted Perturbation Analysis}\label{h:method}

Feature attribution methods for time series classification identify the time points that influence a model's predictions.
Since there is usually no ground truth for evaluating attributions, perturbation analysis is commonly used to assess attribution quality by modifying input features and observing the impact on model predictions.
The assumption is that destroying information at important time points should cause the predictions to change, while perturbing irrelevant time points should have minimal impact.

Let $\mathbf{x} = [x_1, \dots, x_N]$ represent a univariate time series of length $N$. A classifier $f(\mathbf{x})$ outputs predicted probabilities over $C$ classes, where $q_c$ denotes the probability of class $c$. An attribution method produces relevance scores $\mathbf{r} = [r_1, \dots, r_N]$ of the same length as $\mathbf{x}$, where $r_i$ quantifies the importance of time point $i$ to the model's prediction.
To evaluate whether attributions correctly identify relevant features, we perturb the time series $\mathbf{x}$ using a perturbation strategy $p$. The perturbed value at time point $i$ is denoted as $x'_i$.  Common perturbation strategies include replacing values with constants ($x'_i = 0$), statistical aggregates ($x'_i = \text{mean}(\mathbf{x})$), or transformations based on local statistics.

\begin{figure}[!htbp]
\includegraphics[width=\textwidth]{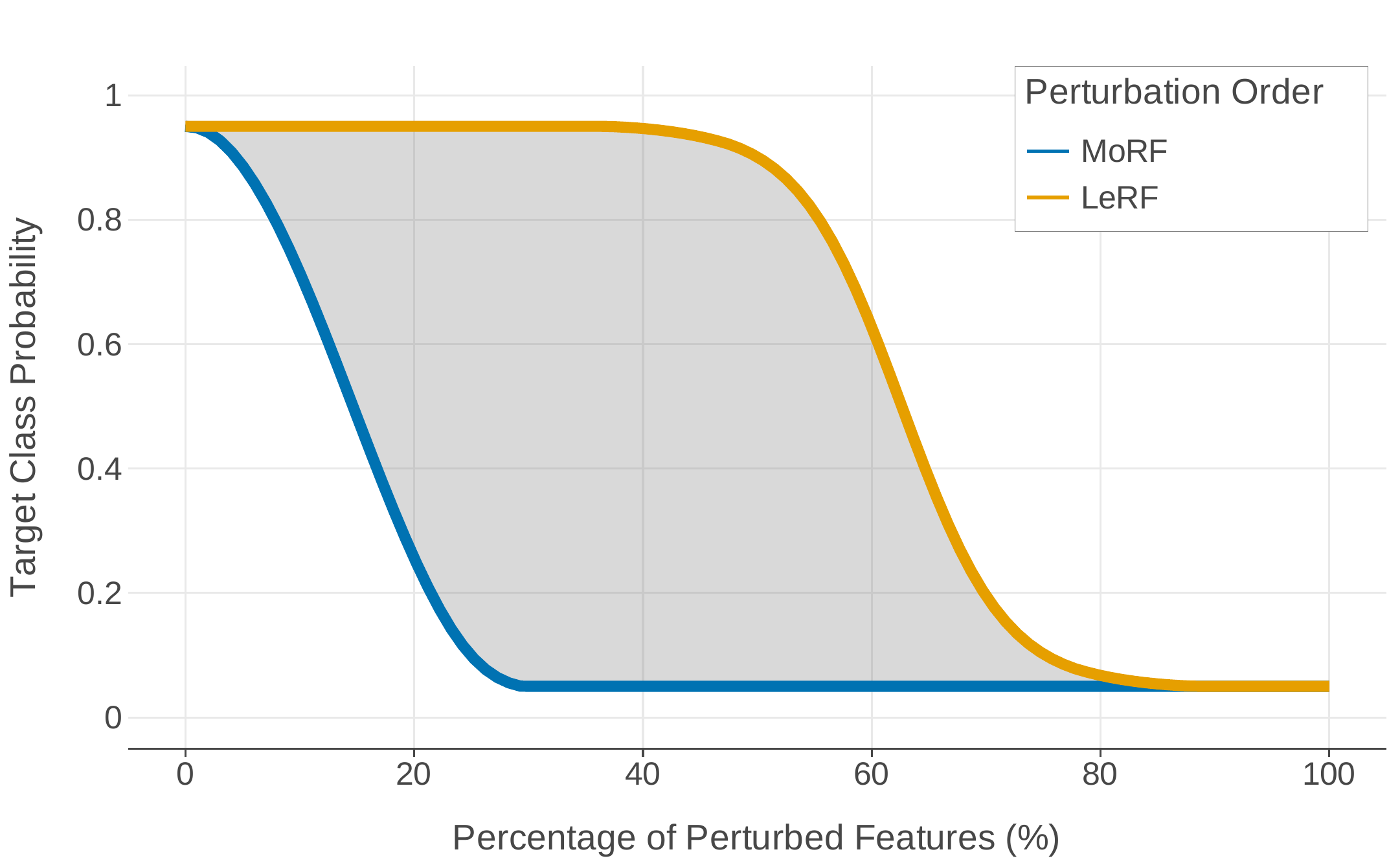}
\caption{Example of MoRF and LeRF perturbation curves. Here, we show a desirable outcome, where $\text{PC}_{\text{MoRF}}$ drops quickly as a result of perturbation whereas $\text{PC}_{\text{LeRF}}$ stays unaffected when perturbing the first 40\% of least important features, resulting in a high DS, or area between both perturbation curves.}\label{fig:ds_method_example}
\end{figure}

We evaluate attribution quality using the degradation score (DS)~\cite{simic.etal_2022_perturbation,schulz.etal_2020_restricting}. This metric compares two perturbation sequences: most relevant features first (MoRF), where features are perturbed in descending order of attributed importance, and least relevant features first (LeRF), where features are perturbed in ascending order. An effective attribution method should show strong prediction changes under MoRF perturbation but minimal impact under LeRF perturbation. This aligns with the intuition that modifying truly important features should significantly disrupt the model's decision-making process, while perturbing irrelevant features should leave the core signal intact. Figure~\ref{fig:ds_method_example} illustrates these expected behaviors.

Formally, for a given instance $\mathbf{x}$ belonging to class $c$, a perturbation strategy $p$, and a number of perturbed features $m$, we track the prediction changes through the perturbation curve, which is obtained from interpolating the vector:
\begin{equation}
    \label{eq:pc}
    \text{PC}_{\mathbf{x}}(c,p,m) =  [q_{cp1}, \dots, q_{cpm}]\,,
\end{equation}
where $q_{cpi}$ represents the predicted probability of class $c$ after perturbing $i$ features. If all time points are perturbed, then $m = N$. Without loss of generality, we denote the vectors $\text{PC}_{\mathbf{x}}(c,p,m)$ describing the perturbation curves in the MoRF and LeRF order as $\text{PC}_{\text{MoRF}}$ and $\text{PC}_{\text{LeRF}}$, respectively. Then, the DS measures the sum of the signed difference between the LeRF and MoRF perturbation curves at each perturbation step:
\begin{equation}
    \label{eq:ds}
    \text{DS} = \frac{1}{m} \sum_i^m (\text{PC}_{\text{LeRF}_i} - \text{PC}_{\text{MoRF}_i})\,.
\end{equation}
The DS ranges from -1 to 1, where positive values suggest correct feature attributions (MoRF perturbations impact predictions more than LeRF), zero indicates non-discriminative attributions (equal impact regardless of perturbation order), and negative values suggest flawed attributions (LeRF perturbations have a stronger impact than MoRF).

To evaluate attribution methods for multiple instances, we usually compute the mean degradation score $\overline{\text{DS}}$. However, this aggregate metric can mask important class-specific behaviors. Therefore, we extend this evaluation framework by introducing the class-adjusted degradation score $\text{DS}_c$ that balances overall attribution accuracy with consistent performance across classes. This metric rewards attribution methods that achieve both high aggregate performance and uniform behavior across different classes. Formally, we define:
\begin{equation}
\label{eq:ds_c}
\text{DS}_c(\alpha) = \overline{\text{DS}} - \alpha \cdot \Delta\,,
\end{equation}
where $\alpha$ is a parameter in $[0,1]$ that controls the penalty strength. 
The penalty term $\Delta$ quantifies the performance differences between classes. For example, for the trivial case of binary classification, it measures the absolute difference in attribution performance between the two classes:
\begin{equation}
\label{eq:penalty_bin}
\Delta = \frac{1}{2} |\overline{\text{DS}}_1 - \overline{\text{DS}}_0|\,.
\end{equation}
For multi-class problems, we extend this concept by computing the mean absolute difference across all possible class pairs:
\begin{equation}
\label{eq:penalty_multi}
\Delta = \frac{1}{2} \frac{1}{\binom{C}{2}} \sum_{i<j} |\overline{\text{DS}}_i - \overline{\text{DS}}_j| = \frac{1}{C(C-1)} \sum_{i<j} |\overline{\text{DS}}_i - \overline{\text{DS}}_j|\,,
\end{equation}
where $i,j$ are class indices and $C$ is the number of classes.
Since the maximum possible value of the mean absolute difference in all pairs is 2, we introduce a factor $\frac{1}{2}$ that normalizes the penalty to the interval [0,1]. This ensures that the class-adjusted degradation score $\text{DS}_c$ remains on the [-1,1] scale, conserving its interpretation. Setting $\alpha=1$ assigns equal importance to overall attribution correctness ($\overline{\text{DS}}$) and consistency between classes ($\Delta$), ensuring that both aspects are weighted equally in the final evaluation metric.
For example, in a binary classification task where $\overline{\text{DS}}_1=1$ and $\overline{\text{DS}}_0=0$, then $\overline{\text{DS}}=0.5$, which can be interpreted as a moderately good degradation score value. However, the MoRF and LeRF curves behave perfectly for all instances of class 1, while for class 0, no differences are observed in the curves on average. Assuming $\alpha=1$, the penalty $\Delta=0.5$ allows us to account for this difference in class behavior, resulting in a more strict class-adjusted degradation score $\text{DS}_c=0$.

Although we apply this penalty framework to the mean DS, it generalizes to any evaluation metric. This allows a systematic investigation of class-dependent effects in attribution evaluation by comparing base metrics against their class-adjusted variants while requiring minimal additional computation, as it utilizes class information already collected during the standard perturbation process. We use this framework to analyze how perturbation-based evaluation methods can exhibit different behaviors across classes.

\section{Experiment Set-up}\label{h:experiment-set-up}

We design our experiments to systematically investigate the effectiveness of attribution methods for time series data, with a particular focus on class-specific differences in perturbation behavior. Our investigation addresses three key aspects: (1) the general effectiveness of attribution methods across architectures and datasets, (2) the presence and characteristics of class-dependent perturbation effects, and (3) the relationship between perturbation strategy selection and these effects.

We use four univariate time series datasets from the UCR Time Series Classification Archive~\cite{dau.etal_2019_ucr}: \textit{FordA}, \textit{FordB}, \textit{ElectricDevices} (ElecDev) and \textit{Wafer}. Table~\ref{tab:dataset_performance} presents the characteristics of these datasets along with the achieved accuracy of the classifiers used in this study.
For visualizations of representative samples and detailed class distributions, see Appendix~\ref{appendix:datasets}.
These datasets are among the largest in the UCR Archive and are commonly used to evaluate feature attributions with perturbations~\cite{schlegel.etal_2019_rigorous,simic.etal_2022_perturbation,mercier.etal_2022_time,schlegel.keim_2023_deep,schlegel.keim_2023_introducing,turbe.etal_2023_evaluation,mercier.etal_2022_time}, making our results comparable to previous work.

\begin{table}[!htbp]
\caption{Dataset characteristics and model performance across architectures. Accuracy metrics (Train, Val, Test) are reported in decimal format.}
\label{tab:dataset_performance}
\setlength{\tabcolsep}{0pt}
\begin{tabular*}{\textwidth}{@{\extracolsep{\fill}} lrrrr ccc ccc @{}}
\toprule
& & & & & \multicolumn{3}{c}{ResNet} & \multicolumn{3}{c}{InceptionTime} \\
\cmidrule(lr){6-8} \cmidrule(lr){9-11}
Dataset & Train & Test & Length & Classes & Train & Val & Test & Train & Val & Test \\
& Size & Size & & & & & & & & \\
\midrule
FordA & 3,601 & 1,320 & 500 & 2 & 0.999 & 0.931 & 0.937 & 1.000 & 0.945 & 0.952 \\
FordB & 3,636 & 810 & 500 & 2 & 1.000 & 0.930 & 0.804 & 0.997 & 0.938 & 0.849 \\
Wafer & 1,000 & 6,164 & 152 & 2 & 1.000 & 1.000 & 0.993 & 1.000 & 1.000 & 0.998 \\
ElecDev & 8,926 & 7,711 & 96 & 7 & 0.981 & 0.913 & 0.716 & 0.987 & 0.895 & 0.702 \\
\bottomrule
\end{tabular*} 
\end{table} 

For model training, we select two distinct and widely-adopted deep learning architectures: ResNet~\cite{wang.etal_2017_time} and InceptionTime~\cite{ismailfawaz.etal_2020_inceptiontime} for their strong performance baselines~\cite{fawaz.etal_2019_deep} and different feature extraction approaches.
Following the predefined UCR splits, we train each model with a batch size of 256 using the AdamW optimizer with cosine annealing learning rate scheduling for up to 500 epochs, implementing early stopping with patience of 25 epochs to ensure stable model convergence. The validation set for early stopping consists of 20\% of randomly selected observations from the train set, stratified by the class label.
As shown in Table~\ref{tab:dataset_performance}, both architectures achieve performance comparable to previous perturbation studies~\cite{schlegel.keim_2023_deep,simic.etal_2022_perturbation} and approach the performance of state-of-the-art deep learning models~\cite{fawaz.etal_2019_deep}. 
Although we observe some performance disparity between training and test sets, indicating potential model capacity for further optimization through hyperparameter tuning, the achieved performance levels are sufficient for our objective of evaluating feature attributions.

We evaluate five widely adopted attribution methods.\footnote{We implement all attribution methods using the TSInterpret package~\cite{hollig.etal_2023_tsinterpret}.} These include four gradient-based methods: Gradients (GR)~\cite{simonyan.etal_2014_deep}, Integrated Gradients (IG)~\cite{sundararajan.etal_2017_axiomatic}, SmoothGrad (SG)~\cite{smilkov.etal_2017_smoothgrad}, and Gradient SHAP (GS)~\cite{lundberg.lee_2017_unified}. We also include a method based on perturbation, Feature Occlusion (FO)~\cite{fong.vedaldi_2017_interpretable}.
While gradient-based methods compute feature importance by analyzing how changes in inputs affect model outputs through gradient calculations, perturbation-based methods systematically modify input features and observe the resulting changes in predictions.
To ensure balanced class representation while maintaining computational feasibility, we compute attributions on a stratified sample of 300 instances per class from the test set.
Attributions always explain the predicted class label.

Our experimental design incorporates six established perturbation strategies from previous work~\cite{simic.etal_2022_perturbation,schlegel.keim_2023_deep} and extends them with a systematic framework of constant-value perturbations. The details of the different strategies are described in Table~\ref{tab:perturbation_strategies}. Although previous studies primarily focus on mean and zero value substitution among other more sophisticated strategies, we also evaluate a comprehensive grid of constant perturbation values ranging between $-2$ and $2$. This extension provides an interpretable baseline for understanding class-dependent perturbation effects, especially given the normalization of the UCR datasets to zero mean and unit standard deviation.

\begin{table}[!htbp]
\caption{Perturbation strategies investigated in our experiments. Each strategy transforms an input time series, represented as vector $\mathbf{x} = [x_1, \dots, x_N]$ where $x_i$ represents the value at time step $i$. The perturbed value is denoted as $x'_i$. For strategies involving subsequence length $k$, we set $k=0.1$ in our experiments, corresponding to 10\% of the time series length.}
\label{tab:perturbation_strategies}
\begin{tabular*}{\textwidth}{@{\extracolsep{\fill}}l>{\raggedright\arraybackslash}p{0.38\textwidth}>{\raggedright\arraybackslash}p{0.42\textwidth}@{}}
\toprule
Strategy & Description & Modification procedure \\
\midrule
Gauss & Random noise from distribution & $x'_i \sim \mathcal{N}(\text{mean}(\mathbf{x}), \text{std}(\mathbf{x}))$ \\
\addlinespace
Unif & Random values within range & $x'_i \sim \mathcal{U}(\min(\mathbf{x}), \max(\mathbf{x}))$ \\
\addlinespace
Opp & Flip sign & $x'_i = -x_i$ \\
\addlinespace
Inv & Invert around maximum & $x'_i = \max(\mathbf{x}) - x_i$ \\
\addlinespace
SubMean & Local subsequence average & $x'_i = \frac{1}{|W_i|}\sum_{j \in W_i} x_j$, where $W_i = \{j : \max(0,i-k+1) \leq j \leq i\}$\\
\addlinespace
Zero & Replace with zero & $x'_i = 0$ \\
\addlinespace
Constant & Replace with predefined values & $x'_i = c$ where $c \in \{-2, -1.5, -1, -0.5, 0, 0.5, 1, 1.5, 2\}$ \\
\bottomrule
\end{tabular*}
\end{table}

We apply these perturbation strategies systematically following the MoRF and LeRF orders, as determined by the feature attributions. For both orders, features are perturbed incrementally in steps of 2\% of the time series length (rounded up) until 50\% perturbation coverage is reached, recording the predicted probabilities at each perturbation step. This bounded perturbation approach ensures computational efficiency by excluding the latter half of features, which typically exhibit minimal discriminative power. 

To analyze both overall attribution quality and potential class-dependent effects, we employ two evaluation approaches. First, we compute the DS (Equation~\ref{eq:ds}) from these perturbation curves, providing a normalized measure between -1 and 1 for each sample that enables consistent comparison across datasets with varying time series lengths and perturbation step sizes. We establish overall performance by averaging DS metrics across all experimental conditions, allowing comparison with previous perturbation studies. To capture class-dependent effects, we extend this analysis using class-adjusted penalties (Equations~\ref{eq:penalty_bin} and~\ref{eq:penalty_multi}) to calculate the class-adjusted metric, $\text{DS}_c$ (Equation~\ref{eq:ds_c}) with $\alpha=1$.

\section{Results and Discussion}\label{h:results-discussion}

\subsection{Evaluating Attribution Quality}

To establish baseline performance and enable comparison with previous work, we begin by evaluating the overall effectiveness of different attribution methods and perturbation strategies. Table~\ref{tab:avg_ds} presents the mean DS metrics aggregated across all experimental conditions. The observed DS ranges align with previous findings by Šimić et al.~\cite{simic.etal_2022_perturbation}, suggesting consistent behavior in different experimental settings.

\begin{table}[!htbp]
\caption{Mean degradation scores ($\overline{\text{DS}}$) for named perturbation strategies across datasets, models, and attribution methods. 
$\overline{\text{DS}}$ measures the average differential impact between perturbing most and least relevant features. 
Positive values indicate correct feature identification, values near zero suggest non-discriminative attributions, and negative values indicate reversed feature importance. 
The largest values for each dataset and model-attribution column pair are highlighted in bold.}
\label{tab:avg_ds}
\setlength{\tabcolsep}{0pt}
\begin{tabular*}{\textwidth}{@{\extracolsep{\fill}} llcccccccccc @{}}
\toprule
 & & \multicolumn{5}{c}{ResNet} & \multicolumn{5}{c}{InceptionTime} \\
\cmidrule(lr){3-7} \cmidrule(lr){8-12}
Dataset & Perturbation & GR & IG & SG & GS & FO & GR & IG & SG & GS & FO \\
\midrule
\multirow[t]{7}{*}{FordA} & Gauss & -.006 & .058 & .021 & -.040 & .088 & \textbf{-.004} & .109 & .012 & -.025 & .141 \\
 & Inv & \textbf{-.002} & .014 & .008 & .012 & .015 & \textbf{-.004} & .016 & .001 & .011 & .018 \\
 & Opp & -.027 & .031 & .011 & .027 & .040 & -.052 & .058 & .018 & .044 & .061 \\
 & SubMean & -.041 & .074 & \textbf{.026} & .071 & .127 & -.092 & .167 & \textbf{.051} & .143 & .198 \\
 & Unif & -.013 & .008 & .004 & .009 & .011 & -.014 & .020 & .010 & .017 & .019 \\
 & Zero & -.041 & \textbf{.075} & .025 & \textbf{.072} & \textbf{.130} & -.096 & \textbf{.173} & \textbf{.051} & \textbf{.149} & \textbf{.205} \\
\cline{1-12}
\multirow[t]{7}{*}{FordB} & Gauss & -.002 & .021 & -.002 & -.026 & .030 & -.029 & \textbf{.130} & \textbf{.014} & -.025 & \textbf{.182} \\
 & Inv & \textbf{.000} & .001 & .000 & .001 & .002 & \textbf{-.007} & .017 & .007 & .011 & .021 \\
 & Opp & -.012 & .020 & .003 & .015 & .025 & -.045 & .035 & -.022 & .026 & .034 \\
 & SubMean & -.022 & \textbf{.041} & .007 & \textbf{.032} & \textbf{.062} & -.071 & .090 & -.070 & \textbf{.081} & .102 \\
 & Unif & -.003 & .003 & .002 & .003 & .007 & -.009 & .008 & .004 & .003 & .007 \\
 & Zero & -.023 & .040 & \textbf{.008} & .031 & .058 & -.072 & .091 & -.069 & \textbf{.081} & .103 \\
\cline{1-12}
\multirow[t]{7}{*}{Wafer} & Gauss & -.060 & -.105 & .006 & .084 & -.164 & .016 & -.098 & \textbf{.119} & \textbf{.177} & -.204 \\
 & Inv & .022 & .065 & .037 & .043 & .139 & -.013 & \textbf{.093} & .026 & .072 & .244 \\
 & Opp & -.020 & .012 & .013 & .011 & .021 & -.009 & .059 & .044 & .047 & .134 \\
 & SubMean & \textbf{.117} & \textbf{.198} & .013 & \textbf{.172} & \textbf{.329} & \textbf{.060} & .054 & .031 & .067 & .058 \\
 & Unif & -.011 & .012 & .035 & .019 & .034 & .030 & .024 & .066 & .023 & .102 \\
 & Zero & -.019 & .064 & \textbf{.045} & .050 & .138 & .024 & .079 & .044 & .050 & \textbf{.278} \\
\cline{1-12}
\multirow[t]{7}{*}{ElecDev} & Gauss & -.004 & -.010 & .009 & .006 & -.008 & .029 & -.001 & .079 & .050 & -.036 \\
 & Inv & .014 & .059 & .080 & .052 & .071 & -.026 & .060 & .010 & .056 & .028 \\
 & Opp & .023 & .180 & .061 & .144 & .220 & .073 & .183 & .087 & .179 & .197 \\
 & SubMean & \textbf{.143} & \textbf{.236} & \textbf{.220} & \textbf{.222} & \textbf{.272} & \textbf{.148} & \textbf{.217} & \textbf{.113} & \textbf{.201} & \textbf{.290} \\
 & Unif & .052 & .094 & .052 & .057 & .156 & .038 & .103 & .048 & .070 & .107 \\
 & Zero & .035 & .156 & .075 & .122 & .271 & .085 & .187 & .076 & .164 & .269 \\
\cline{1-12}
\bottomrule
\end{tabular*}
\end{table}

Examining these results in detail, we find that the perturbation strategies exhibit varying levels of performance, with notable dependencies on model architectures and datasets. Zero and SubMean perturbations often achieve the highest DS scores across the different experimental configurations, though Gauss occasionally outperforms them in specific contexts. Importantly, optimal perturbation selection appears highly contingent on model architecture, even when controlling for the dataset. For instance, on the FordB and Wafer datasets, Gauss perturbations perform relatively well for InceptionTime on several attribution methods, but for ResNet, SubMean or Zero yield better results.

We also observe substantial performance variations among the attribution methods. FO, IG and GS tend to achieve the highest DS values across datasets and perturbation strategies, indicating better feature importance identification. While FO frequently demonstrates marginally superior performance compared to IG and GS, this advantage is not consistent across all experimental settings. In contrast, GR shows poor discriminative ability, frequently yielding negative or near-zero DS values, suggesting that its feature importance assignments are often not better than random ordering. Between these extremes, GS typically outperforms GR but falls short of the effectiveness demonstrated by IG, GS and FO, depending on the experimental condition.

Although these aggregate metrics provide valuable insights into overall method effectiveness, they potentially mask important variations in performance distributions. To better understand these underlying patterns, we examine the distribution of DS metrics through a more detailed case study.

\subsection{Distribution Patterns in Attribution Quality}

\begin{figure}[!htbp]
    \centering
    \begin{subfigure}[b]{\textwidth}
        \centering
        \includegraphics[width=\textwidth]{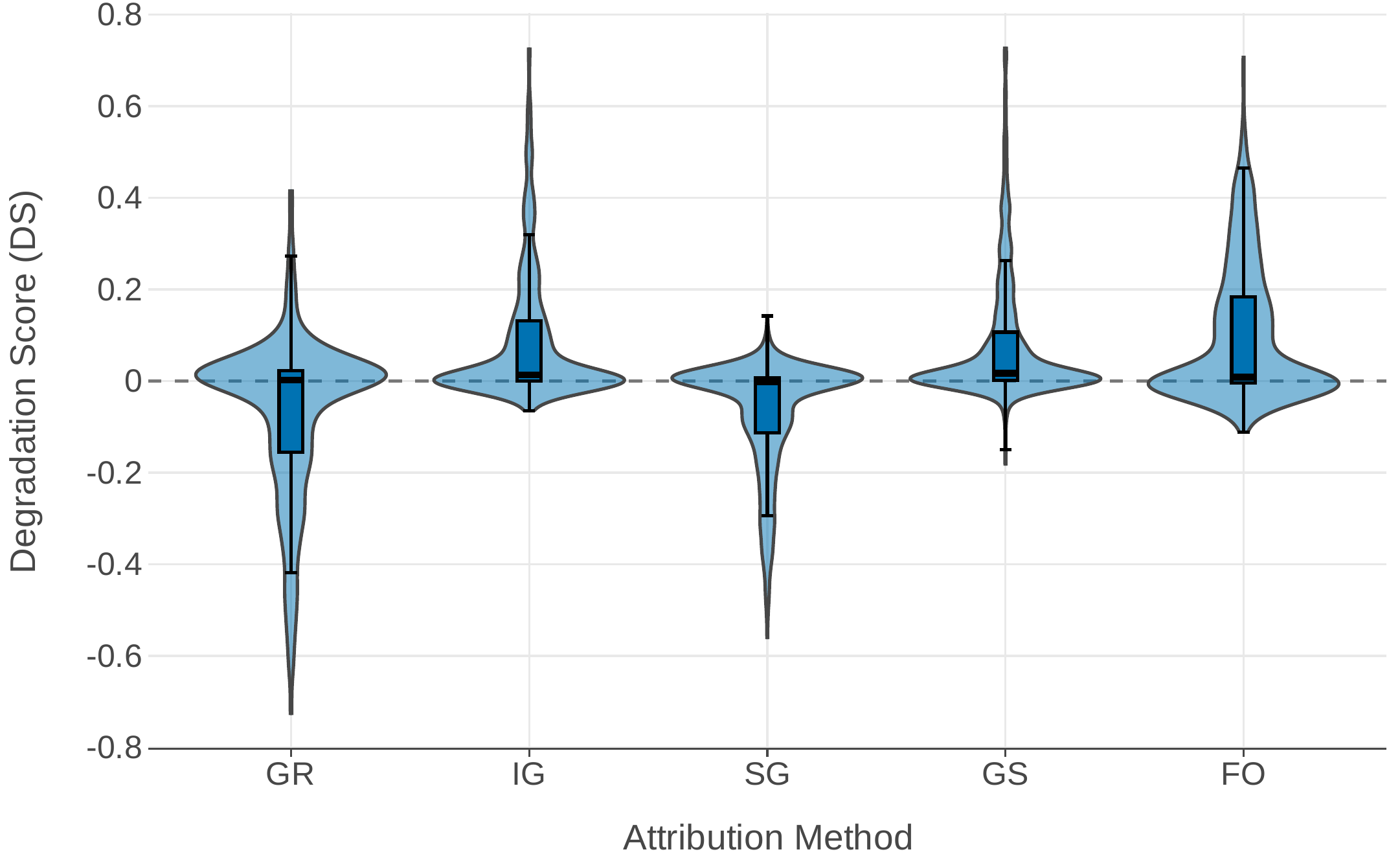}
        \caption{Overall DS distributions per attribution method}
        \label{fig:ds_distributions_a}
    \end{subfigure}
    \vspace{0.8em}
    \begin{subfigure}[b]{\textwidth}
        \centering
        \includegraphics[width=\textwidth]{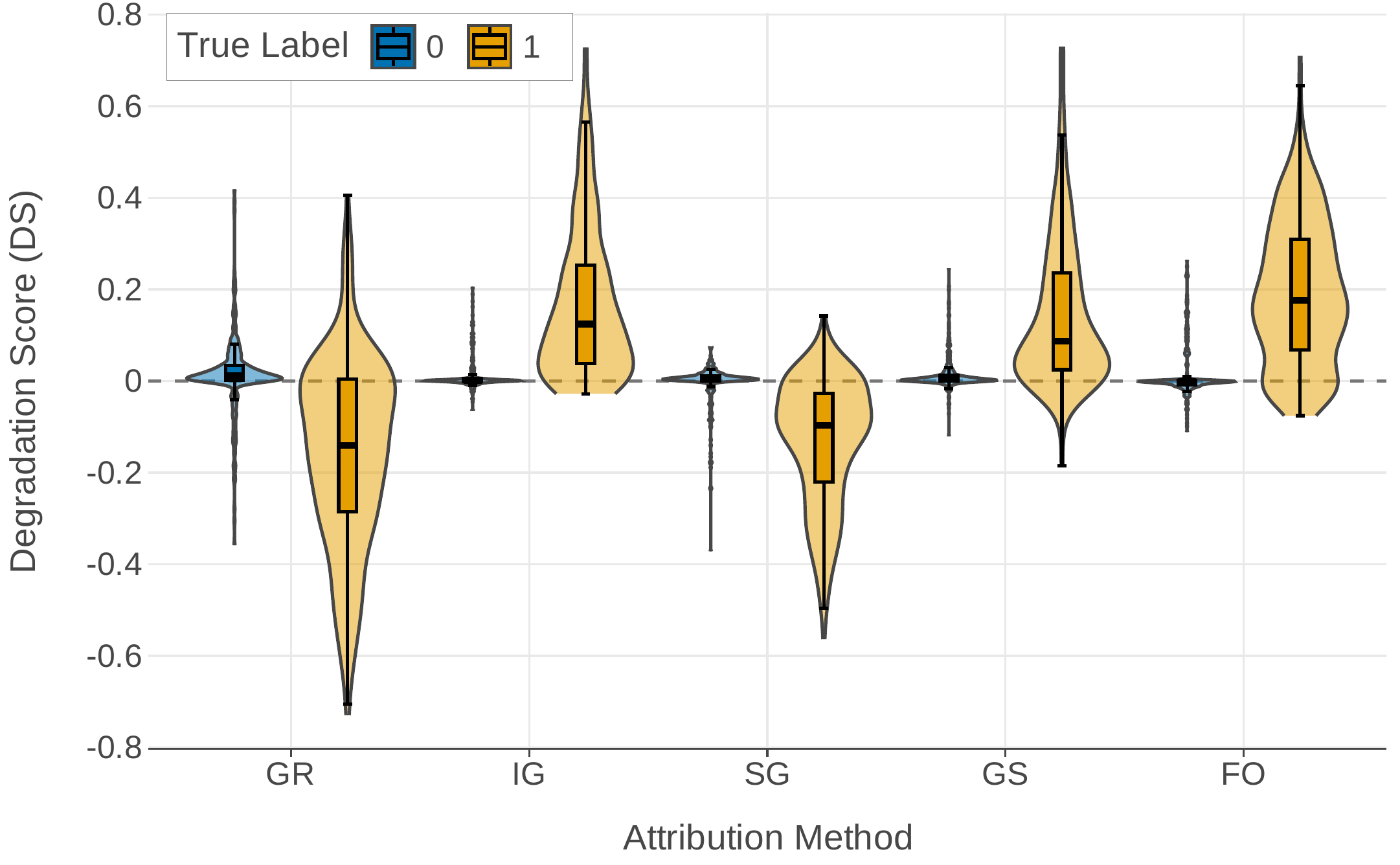}
        \caption{Class-specific DS distributions per attribution method}
        \label{fig:ds_distributions_b}
    \end{subfigure}
    \caption{Distributions of DS for different attribution methods on the FordB dataset (classifier: InceptionTime, perturbation strategy: SubMean).}
    \label{fig:ds_distributions}
\end{figure}

To understand how attribution performance varies between individual instances, we analyze the distributional characteristics of DS scores. For this analysis, we focus on the FordB dataset, InceptionTime architecture, and SubMean perturbation strategy, as these demonstrate patterns typical of our broader findings. Figure~\ref{fig:ds_distributions} presents these distributions, revealing both overall performance patterns and, crucially, class-specific effects.

The aggregate distributions shown in Figure~\ref{fig:ds_distributions_a} reveal notable variation in the effectiveness of attribution methods. IG, GS, and FO exhibit positively skewed distributions with extended tails toward higher DS values, indicating better attribution quality. In contrast, GR and SG demonstrate negative skewness with tails extending toward lower DS values, suggesting less reliable feature identification. Critically, we observe a concentration of scores around zero, implying limited discriminative power in feature ordering for many instances, even among the better-performing methods.

Figure~\ref{fig:ds_distributions_b} presents a class-stratified analysis that uncovers substantial heterogeneity in attribution quality across classes.  For IG, GS, and FO, instances from class 1 consistently achieve higher DS scores with pronounced positive tails and minimal negative values. The same methods show markedly different behavior for class 0, where DS scores cluster tightly around zero, indicating minimal perturbation impact. GR and SG show an inverse pattern, with class 1 showing predominantly negative scores while class 0 maintains a more balanced distribution centered near zero.

These pronounced class-dependent variations in attribution performance raise questions about the generalizability of perturbation-based evaluation methods.  To determine whether this phenomenon extends beyond our case study, we now examine class-dependent behavior across all experimental conditions.

\subsection{Class-Dependent Effects in Perturbation Analysis}

We apply the proposed class-adjusted DS ($\text{DS}_c$) metric to all experimental conditions to investigate class-dependent behaviors. Table~\ref{tab:adjusted_ds} presents these adjusted scores, which balance average attribution correctness with consistency between classes. Our results show that incorporating class consistency penalties substantially reduces performance metrics across most experimental conditions, particularly for previously high-performing perturbation strategies.

\begin{table}[!htbp]
\caption{Class-adjusted mean degradation scores ($\text{DS}_c$ with $\alpha=1$) for named perturbation strategies across datasets, models, and attribution methods.
$\text{DS}_c$ measures both attribution quality and cross-class consistency of perturbations.
Positive values indicate good attribution quality with consistent behavior across classes, while lower values suggest either poor attribution quality or inconsistent performance between classes.
The largest values for each dataset and model-attribution column pair are highlighted in bold.}
\label{tab:adjusted_ds}
\setlength{\tabcolsep}{0pt}
\begin{tabular*}{\textwidth}{@{\extracolsep{\fill}} llcccccccccc @{}}
\toprule
 & & \multicolumn{5}{c}{ResNet} & \multicolumn{5}{c}{InceptionTime} \\
\cmidrule(lr){3-7} \cmidrule(lr){8-12}
Dataset & Perturbation & GR & IG & SG & GS & FO & GR & IG & SG & GS & FO \\
\midrule
\multirow[t]{7}{*}{FordA} & Gauss & -.011 & \textbf{.003} & \textbf{.001} & -.078 & .003 & -.008 & \textbf{.001} & .000 & -.049 & \textbf{.002} \\
 & Inv & \textbf{-.005} & .000 & .000 & .000 & .001 & \textbf{-.007} & .000 & .000 & \textbf{.000} & .000 \\
 & Opp & -.053 & .001 & .000 & .001 & .001 & -.104 & .000 & .000 & \textbf{.000} & .000 \\
 & SubMean & -.082 & \textbf{.003} & .000 & \textbf{.002} & \textbf{.005} & -.188 & .000 & \textbf{.001} & \textbf{.000} & .000 \\
 & Unif & -.025 & .000 & .000 & .000 & .000 & -.028 & .000 & .000 & \textbf{.000} & .000 \\
 & Zero & -.082 & \textbf{.003} & \textbf{.001} & \textbf{.002} & \textbf{.005} & -.197 & .000 & \textbf{.001} & \textbf{.000} & -.001 \\
\cline{1-12}
\multirow[t]{7}{*}{FordB} & Gauss & -.004 & .003 & -.003 & -.049 & .006 & -.068 & \textbf{.015} & \textbf{.009} & -.046 & \textbf{.026} \\
 & Inv & \textbf{-.001} & .000 & -.001 & .000 & .000 & \textbf{-.013} & .003 & .001 & .002 & .003 \\
 & Opp & -.023 & .003 & \textbf{.001} & .002 & .004 & -.091 & .004 & -.041 & .004 & .005 \\
 & SubMean & -.044 & \textbf{.006} & .000 & \textbf{.005} & \textbf{.010} & -.158 & .013 & -.134 & .014 & .014 \\
 & Unif & -.007 & .001 & .000 & .001 & .001 & -.017 & .001 & .001 & .001 & .001 \\
 & Zero & -.044 & \textbf{.006} & .000 & \textbf{.005} & \textbf{.010} & -.159 & .013 & -.134 & \textbf{.015} & .015 \\
\cline{1-12}
\multirow[t]{7}{*}{Wafer} & Gauss & -.149 & -.214 & \textbf{.001} & .003 & -.367 & .003 & -.201 & -.004 & .033 & -.419 \\
 & Inv & .008 & .003 & .000 & .002 & .002 & -.026 & .001 & -.012 & .008 & .009 \\
 & Opp & -.042 & .002 & -.001 & .003 & .003 & -.021 & .002 & -.011 & .010 & .008 \\
 & SubMean & \textbf{.034} & \textbf{.018} & -.005 & \textbf{.037} & \textbf{.038} & \textbf{.014} & \textbf{.048} & \textbf{.000} & .018 & \textbf{.028} \\
 & Unif & -.020 & .001 & \textbf{.001} & .002 & .001 & .002 & .002 & -.014 & .008 & .008 \\
 & Zero & -.047 & .007 & -.001 & .011 & .013 & .012 & .009 & -.023 & \textbf{.036} & .016 \\
\cline{1-12}
\multirow[t]{7}{*}{ElecDev} & Gauss & -.039 & -.036 & -.020 & -.029 & -.044 & .011 & -.015 & .045 & .027 & -.091 \\
 & Inv & -.024 & .012 & .047 & .009 & .025 & -.082 & .018 & -.045 & .027 & -.036 \\
 & Opp & -.029 & .104 & .001 & .096 & .156 & .039 & .140 & .041 & .116 & .122 \\
 & SubMean & \textbf{.078} & \textbf{.130} & \textbf{.141} & \textbf{.136} & .219 & \textbf{.075} & \textbf{.151} & \textbf{.081} & \textbf{.134} & \textbf{.229} \\
 & Unif & .026 & .031 & -.003 & .020 & .082 & -.004 & .055 & .026 & .041 & .022 \\
 & Zero & -.010 & .089 & .031 & .063 & \textbf{.226} & .053 & .118 & .038 & .104 & .162 \\
\cline{1-12}
\bottomrule
\end{tabular*}
\end{table}

Zero, SubMean and Gauss perturbations, which demonstrated superior performance before, show marked degradation under class-adjusted evaluation, with most DS scores shifting toward zero or even becoming negative.
The extent of this impact differs among experimental conditions, particularly across various datasets. Wafer and ElecDev generally show more resilience to class adjustment compared to FordA and FordB, though we still see reduced scores. For ElecDev, the interpretation of these results requires additional context: as the only multiclass dataset in our evaluation, the pairwise averaging of class differences may underestimate class-specific effects due to the higher dimensionality of the classification space.

\begin{figure}[!htbp]
    \centering
    \begin{subfigure}[b]{\textwidth}
        \centering
        \includegraphics[width=\textwidth]{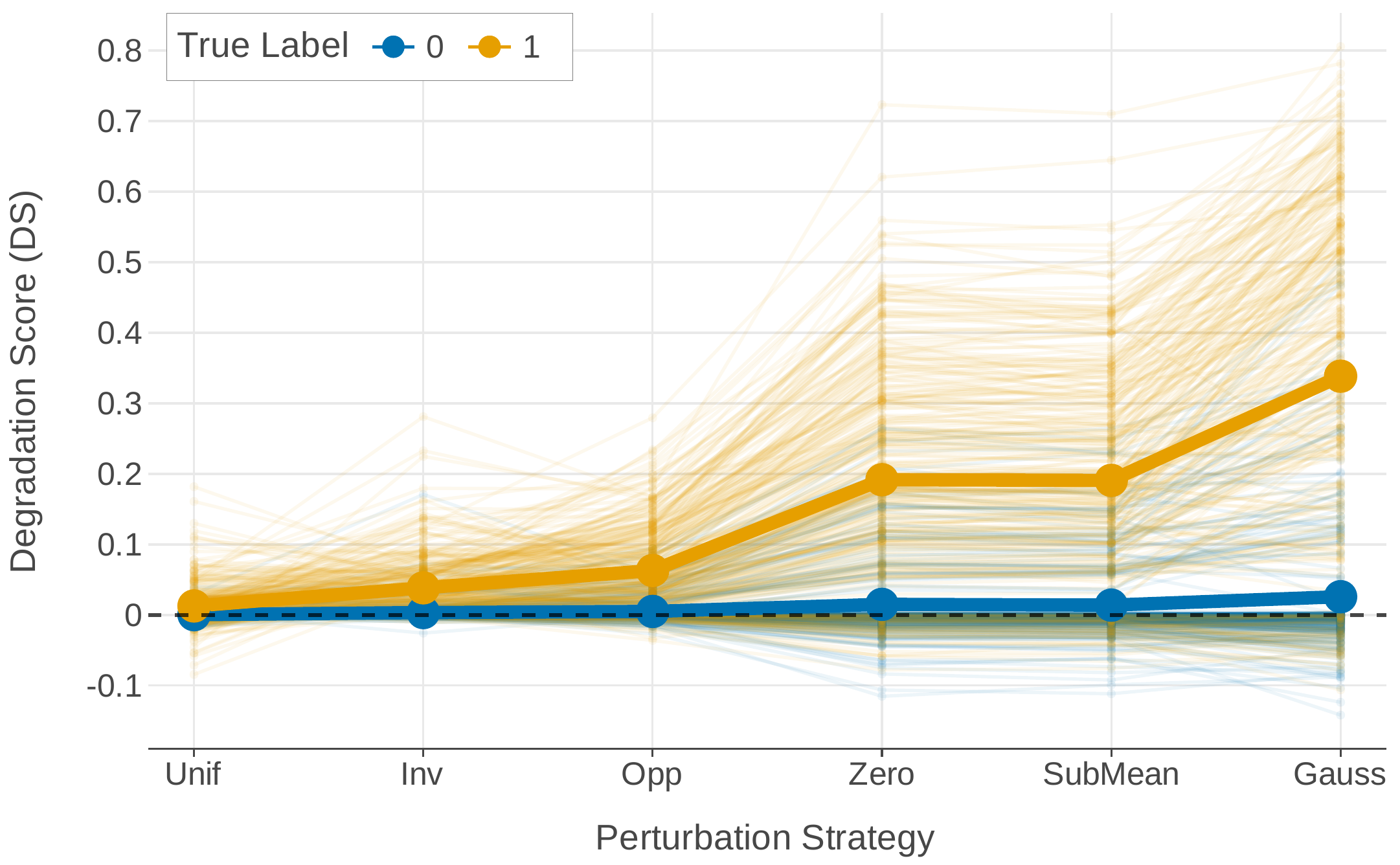}
        \caption{DS for named perturbation strategies}
        \label{fig:perturbations_by_class_a}
    \end{subfigure}
    \vspace{0.8em}
    \begin{subfigure}[b]{\textwidth}
        \centering
        \includegraphics[width=\textwidth]{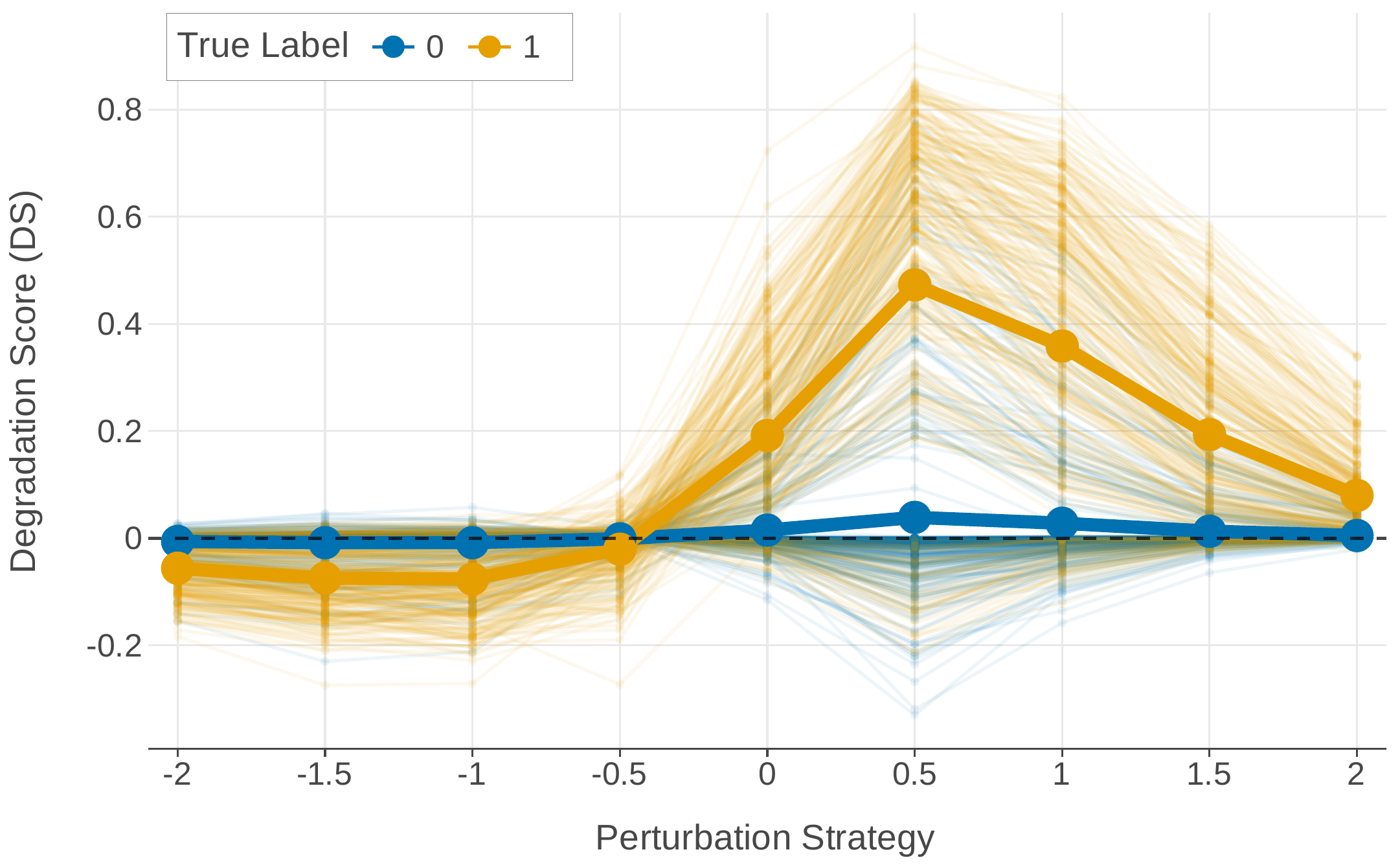}
        \caption{DS for constant perturbation strategies}
        \label{fig:perturbations_by_class_b}
    \end{subfigure}
    \caption{Class-stratified analysis of DS for different perturbation strategies on the FordB dataset (classifier: InceptionTime, attribution method: FO). Thin lines show individual observations, while thick lines indicate class means.}
    \label{fig:perturbations_by_class}
\end{figure}

To examine the mechanisms underlying these class-specific effects, Figure~\ref{fig:perturbations_by_class} presents a detailed analysis of perturbation impacts across classes on the FordB dataset. Figure~\ref{fig:perturbations_by_class_a}, which focuses on named perturbation strategies with InceptionTime and FO attribution, reveals systematic class-dependent behavior. 
Class 0 instances exhibit minimal response to perturbation across all strategies, with DS scores tightly clustered around zero, while Class 1 instances demonstrate substantial variability in perturbation response. For Class 1, Gauss is the most effective, followed by Zero and SubMean strategies.
This asymmetric response pattern may explain the earlier observed degradation in class-adjusted metrics: perturbation strategies succeed primarily by exploiting class-specific model behaviors.

Analyzing the constant-value perturbations in Figure~\ref{fig:perturbations_by_class_b} reinforces these findings while providing additional information. The evaluation of perturbation values between -2 and 2 reveals optimal effectiveness at moderate positive values, particularly around 0.5. We now extend the investigation of the constant perturbation strategies to all experimental conditions.

\subsection{Evaluating Constant Perturbation Values}

Table~\ref{tab:adjusted_ds_constant} shows the class-adjusted DS for the constant perturbation strategies, revealing significant domain-specific variations. The effectiveness of constant perturbations varies significantly by dataset and classifier architecture. FordA and FordB often show the best performance with 0.5 perturbation. Wafer demonstrates inconsistent responses, where optimal perturbation constants vary across attribution methods. ElecDev exhibits the best performance with a perturbation constant of 0.5 for ResNet, and different constants between -0.5 and 0.5 for InceptionTime, though its multiclass nature introduces additional complexity in interpreting class-adjusted metrics. 

Interestingly, negative perturbation values generally yield negative class-adjusted scores for most datasets, indicating asymmetric class responses to the perturbation direction. However, this effect varies by attribution method, classifier, and dataset. For example, GR, SG and GS on ElecDev exclusively show positive scores for negative perturbation constants. These results suggest that the effectiveness of perturbation-based evaluation depends on the alignment between perturbation values and the learned class representations in the feature space.

\begin{table}[!htbp]
\caption{Class-adjusted mean degradation scores ($\text{DS}_c$ with $\alpha=1$) for constant-value perturbation strategies across datasets, models, and attribution methods.
$\text{DS}_c$ measures both attribution quality and cross-class consistency of perturbations.
Positive values indicate good attribution quality with consistent behavior across classes, while lower values suggest either poor attribution quality or inconsistent performance between classes.
The largest values for each dataset and model-attribution column pair are highlighted in bold.}
\label{tab:adjusted_ds_constant}
\setlength{\tabcolsep}{0pt}
\begin{tabular*}{\textwidth}{@{\extracolsep{\fill}} llcccccccccc @{}}
\toprule
 & & \multicolumn{5}{c}{ResNet} & \multicolumn{5}{c}{InceptionTime} \\
\cmidrule(lr){3-7} \cmidrule(lr){8-12}
Dataset & Perturbation & GR & IG & SG & GS & FO & GR & IG & SG & GS & FO \\
\midrule
\multirow[t]{7}{*}{FordA} & -2 & -.045 & -.068 & -.008 & -.067 & -.083 & -.016 & -.077 & .000 & -.042 & -.083 \\
 & -1.5 & -.066 & -.093 & -.012 & -.092 & -.094 & -.027 & -.100 & .000 & -.056 & -.106 \\
 & -1 & -.098 & -.102 & -.013 & -.104 & -.086 & -.054 & -.091 & .000 & -.054 & -.091 \\
 & -0.5 & -.129 & -.011 & .000 & -.016 & .001 & -.136 & .000 & \textbf{.001} & .000 & .000 \\
 & 0 & -.082 & .003 & \textbf{.001} & .002 & .005 & -.197 & .000 & \textbf{.001} & .000 & -.001 \\
 & 0.5 & -.032 & \textbf{.005} & \textbf{.001} & \textbf{.004} & \textbf{.010} & -.040 & \textbf{.001} & .000 & \textbf{.001} & \textbf{.002} \\
 & 1 & -.012 & .003 & \textbf{.001} & .003 & .004 & -.009 & \textbf{.001} & .000 & .000 & \textbf{.002} \\
 & 1.5 & \textbf{-.009} & .001 & \textbf{.001} & .001 & .001 & -.005 & \textbf{.001} & .000 & .000 & .001 \\
 & 2 & \textbf{-.009} & .000 & .000 & .000 & .001 & \textbf{-.004} & .000 & .000 & .000 & .000 \\
\cline{1-12}
\multirow[t]{7}{*}{FordB} & -2 & -.011 & -.038 & \textbf{.001} & -.043 & -.015 & \textbf{-.004} & -.045 & .000 & -.039 & -.056 \\
 & -1.5 & -.017 & -.045 & \textbf{.001} & -.054 & -.018 & -.007 & -.060 & -.002 & -.054 & -.074 \\
 & -1 & -.026 & -.033 & \textbf{.001} & -.043 & -.009 & -.017 & -.063 & -.017 & -.060 & -.076 \\
 & -0.5 & -.040 & .000 & \textbf{.001} & .000 & .003 & -.054 & -.016 & -.062 & -.023 & -.020 \\
 & 0 & -.044 & .006 & .000 & .005 & .010 & -.159 & .013 & -.134 & .015 & .015 \\
 & 0.5 & -.024 & \textbf{.008} & .000 & \textbf{.007} & \textbf{.018} & -.096 & \textbf{.022} & -.021 & \textbf{.020} & \textbf{.039} \\
 & 1 & -.005 & .003 & -.003 & .003 & .006 & -.069 & .015 & \textbf{.009} & .013 & .028 \\
 & 1.5 & \textbf{-.001} & .001 & -.002 & .001 & .002 & -.053 & .009 & .006 & .008 & .013 \\
 & 2 & \textbf{-.001} & .000 & -.001 & .000 & .000 & -.034 & .005 & .003 & .004 & .005 \\
\cline{1-12}
\multirow[t]{7}{*}{Wafer} & -2 & -.035 & -.054 & \textbf{.001} & -.053 & -.125 & .001 & -.116 & \textbf{-.002} & -.155 & -.374 \\
 & -1.5 & -.087 & -.166 & \textbf{.001} & -.191 & -.289 & .002 & -.169 & -.004 & -.250 & -.423 \\
 & -1 & -.204 & -.202 & .000 & -.267 & -.309 & .006 & -.226 & -.006 & -.314 & -.337 \\
 & -0.5 & -.142 & -.048 & -.004 & -.075 & -.048 & \textbf{.018} & -.177 & -.006 & -.224 & .018 \\
 & 0 & -.047 & .007 & -.001 & \textbf{.011} & .013 & .012 & .009 & -.023 & .036 & .016 \\
 & 0.5 & .008 & \textbf{.008} & .000 & .008 & \textbf{.017} & .005 & \textbf{.014} & -.029 & \textbf{.047} & .047 \\
 & 1 & .025 & .002 & \textbf{.001} & .003 & .016 & -.014 & .004 & -.027 & .031 & \textbf{.056} \\
 & 1.5 & \textbf{.031} & -.003 & -.001 & -.001 & .006 & .000 & .003 & -.016 & .017 & .040 \\
 & 2 & .019 & -.006 & -.003 & -.003 & -.001 & -.001 & .003 & -.013 & .014 & .030 \\
\cline{1-12}
\multirow[t]{7}{*}{ElecDev} & -2 & -.055 & -.040 & -.032 & -.037 & -.054 & .004 & -.028 & .044 & .007 & -.114 \\
 & -1.5 & -.023 & -.039 & -.010 & -.031 & -.029 & .016 & -.010 & .048 & .024 & -.074 \\
 & -1 & \textbf{.017} & -.069 & .000 & -.027 & -.005 & .021 & .003 & .054 & .036 & -.021 \\
 & -0.5 & .002 & -.028 & .018 & .011 & .049 & \textbf{.054} & .055 & \textbf{.075} & .092 & .104 \\
 & 0 & -.010 & .089 & .031 & .063 & .226 & .053 & \textbf{.118} & .038 & .104 & \textbf{.162} \\
 & 0.5 & .005 & \textbf{.147} & \textbf{.100} & \textbf{.129} & \textbf{.269} & .013 & .106 & .019 & \textbf{.106} & .144 \\
 & 1 & .006 & .114 & .076 & .098 & .193 & .005 & .112 & .027 & .094 & .123 \\
 & 1.5 & .005 & .087 & .078 & .086 & .139 & .015 & .098 & .032 & .066 & .063 \\
 & 2 & .003 & .067 & .078 & .068 & .078 & .024 & .041 & .036 & .029 & -.005 \\
\cline{1-12}
\bottomrule
\end{tabular*}
\end{table}

\subsection{Synthesis and Implications}

Our investigation reveals three key findings on perturbation-based evaluation of feature attributions. First, the effectiveness of perturbation strategies can vary substantially between classes, with performance often showing asymmetric impacts. This pattern appears particularly in binary classification tasks, where perturbation strategies may validate attributions effectively for one class while showing limited sensitivity to the other.

Second, we observe that perturbation strategies that show strong aggregate performance often demonstrate the most pronounced class-dependent effects. When examining these strategies using our class-adjusted framework, we find that much of their effectiveness stems from strong performance in specific classes rather than consistent behavior across all classes. This observation suggests the importance of examining class-specific responses when evaluating perturbation strategies.

Third, our analysis of constant perturbation strategies shows that optimal perturbation values can vary across datasets and architectures, indicating that perturbation effectiveness may be influenced by the specific characteristics of learned model representations. This finding suggests that the choice of perturbation values warrants careful consideration for each specific application context.

\section{Conclusion and Future Work}\label{h:conclusion}

This paper presents a systematic investigation of class-dependent effects in perturbation-based evaluation of feature attributions for time series classification. Through empirical evaluation across four datasets, five attribution methods, and multiple perturbation strategies, we demonstrate that perturbation-based evaluation methods can exhibit class-specific behaviors that warrant careful consideration in validation procedures.
Namely, we show that: (1) perturbation effectiveness can vary substantially between classes; (2) strategies with stronger aggregate performance often exhibit more pronounced class-dependent effects; and (3) optimal perturbation strategies can vary considerably across datasets and model architectures, suggesting that perturbation effectiveness is influenced by specific characteristics of learned model representations and that domain-specific calibration may be necessary.

We recommend several evaluation approaches to address potential class-dependent effects. First, supplementing aggregate metrics with more detailed class-stratified analysis can help identify when dominant classes disproportionately influence results. Second, applying penalty frameworks like the one proposed in this study quantifies class-dependent effects without additional computational burden, allowing researchers to systematically compare attribution methods while accounting for class biases. Third, evaluating individual instances with multiple perturbation strategies targeting different class predictions may help validate attribution robustness beyond the limitations of any single approach.

The generalizability of our findings faces two primary constraints. Our investigation encompasses only four datasets and two classifier architectures, potentially limiting broader applicability. Additionally, the observed perturbation response patterns may reflect specific characteristics of our experimental configuration rather than fundamental properties of the evaluation approach. Nevertheless, the alignment with previous literature suggests wider relevance of our methodology.

Future research directions emerge from these findings. One promising avenue involves developing perturbation strategies that systematically account for class-specific model behaviors, particularly focusing on methods that can push predictions toward different classes. Additionally, investigating methods to adaptively select perturbation strategies for individual instances could improve evaluation effectiveness, as our results suggest that different instances may require different perturbation approaches to effectively validate their attributions.

\appendix

\section{Dataset Visualizations and Class Distributions} \label{appendix:datasets} 
\begin{table}[!htbp]
\caption{Class distribution across datasets. Values represent the proportion of samples in each class for training and test sets.}
\label{tab:class_distribution}
\setlength{\tabcolsep}{0pt}
\begin{tabular*}{\textwidth}{@{\extracolsep{\fill}} l rrrrrrrr rrrrrrrr}
\toprule
& \multicolumn{7}{c}{Train} & & \multicolumn{7}{c}{Test} \\
\cmidrule(lr){2-8} \cmidrule(lr){10-16}
Dataset & C0 & C1 & C2 & C3 & C4 & C5 & C6 & & C0 & C1 & C2 & C3 & C4 & C5 & C6 \\
\midrule
FordA   & .513 & .487 & -- & -- & -- & -- & -- & & .516 & .484 & -- & -- & -- & -- & -- \\
FordB   & .512 & .488 & -- & -- & -- & -- & -- & & .495 & .505 & -- & -- & -- & -- & -- \\
Wafer   & .097 & .903 & -- & -- & -- & -- & -- & & .108 & .892 & -- & -- & -- & -- & -- \\
ElecDev & .081 & .250 & .095 & .165 & .270 & .057 & .082 & & .087 & .254 & .098 & .151 & .242 & .096 & .072 \\
\bottomrule
\end{tabular*} 
\end{table}
\begin{figure}[!htbp]
    \centering
    \begin{subfigure}[b]{\textwidth}
        \centering
        \includegraphics[width=\textwidth]{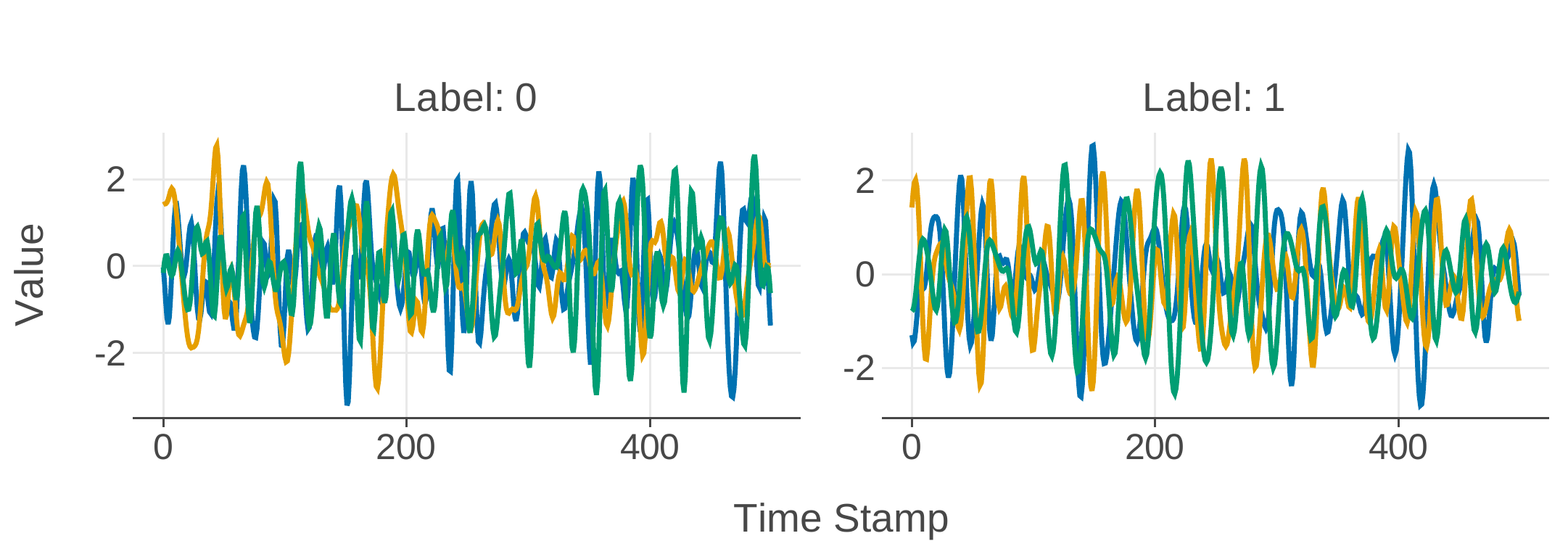}
        \caption{FordA dataset}
        \label{fig:samples_forda}
    \end{subfigure}
    \vspace{0.8em}
    \begin{subfigure}[b]{\textwidth}
        \centering
        \includegraphics[width=\textwidth]{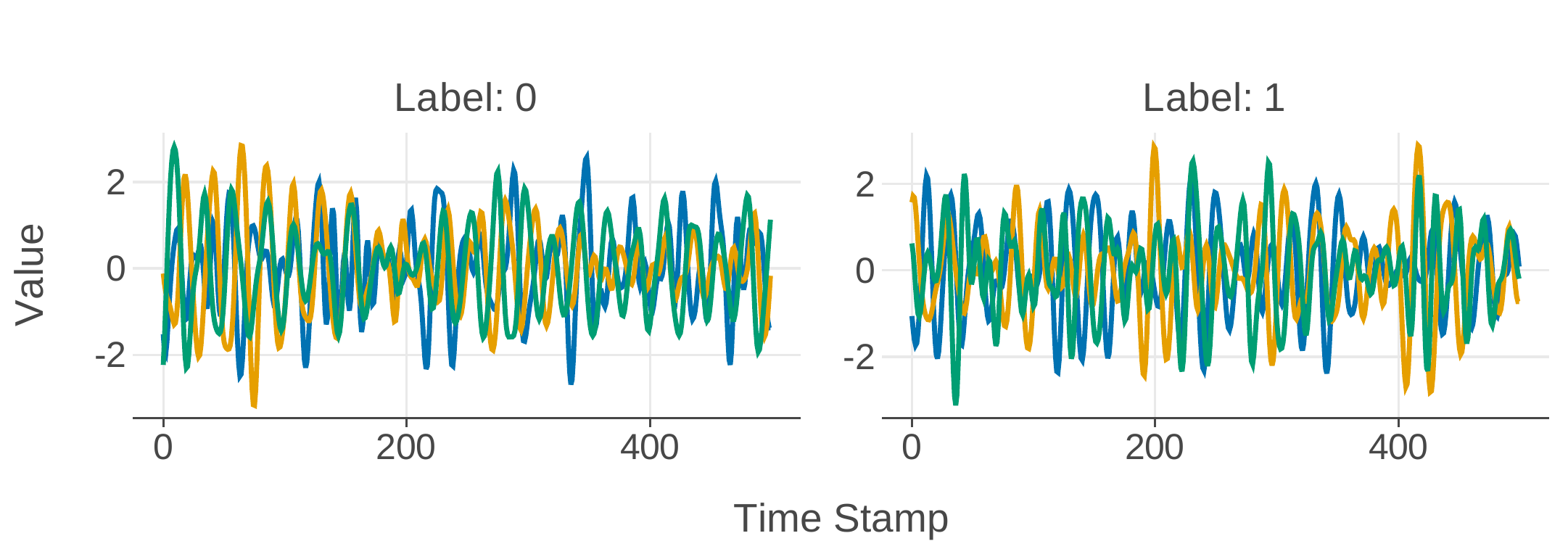}
        \caption{FordB dataset}
        \label{fig:samples_fordb}
    \end{subfigure}
    \vspace{0.8em}
    \begin{subfigure}[b]{\textwidth}
        \centering
        \includegraphics[width=\textwidth]{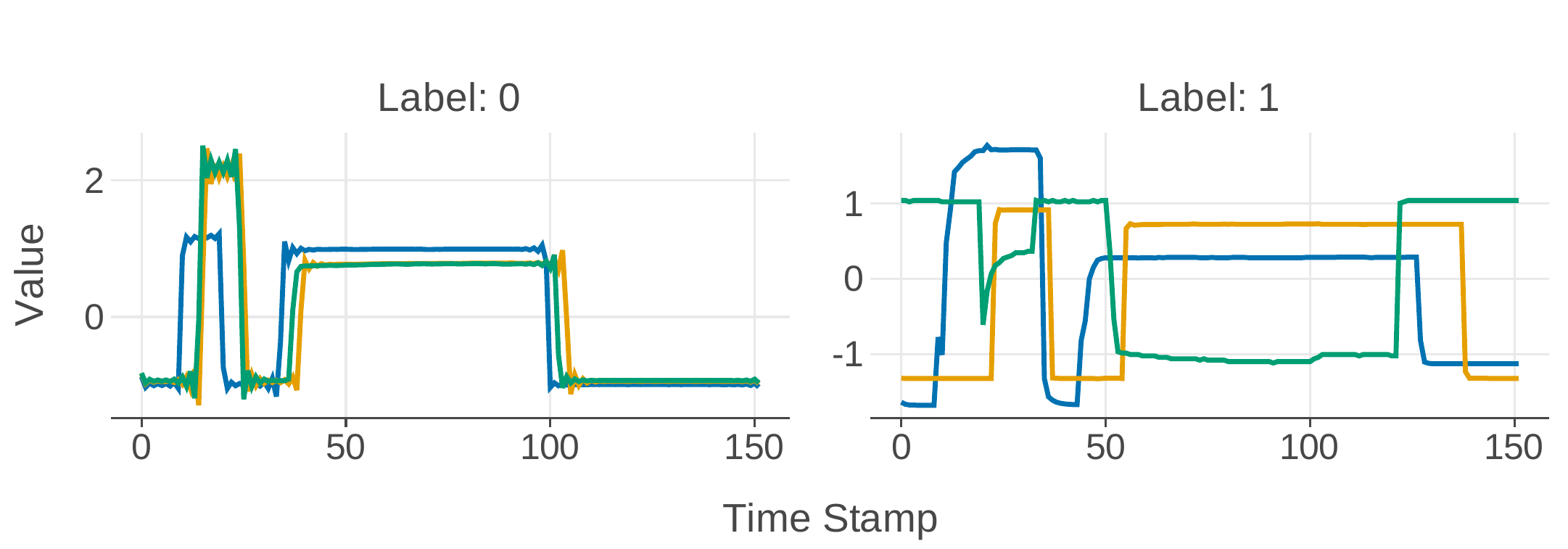}
        \caption{Wafer dataset}
        \label{fig:samples_wafer}
    \end{subfigure}
    \caption{Representative samples from the three binary classification datasets used in this study. For each dataset, three randomly selected instances per class are shown to illustrate the characteristic patterns and variability between classes. If datasets had labels not starting from 0, they were remapped to start from 0 (ranging from 0 to $C-1$), preserving the existing order.}
    \label{fig:samples_binary}
\end{figure}
\begin{figure}[!htbp]
\includegraphics[width=\textwidth]{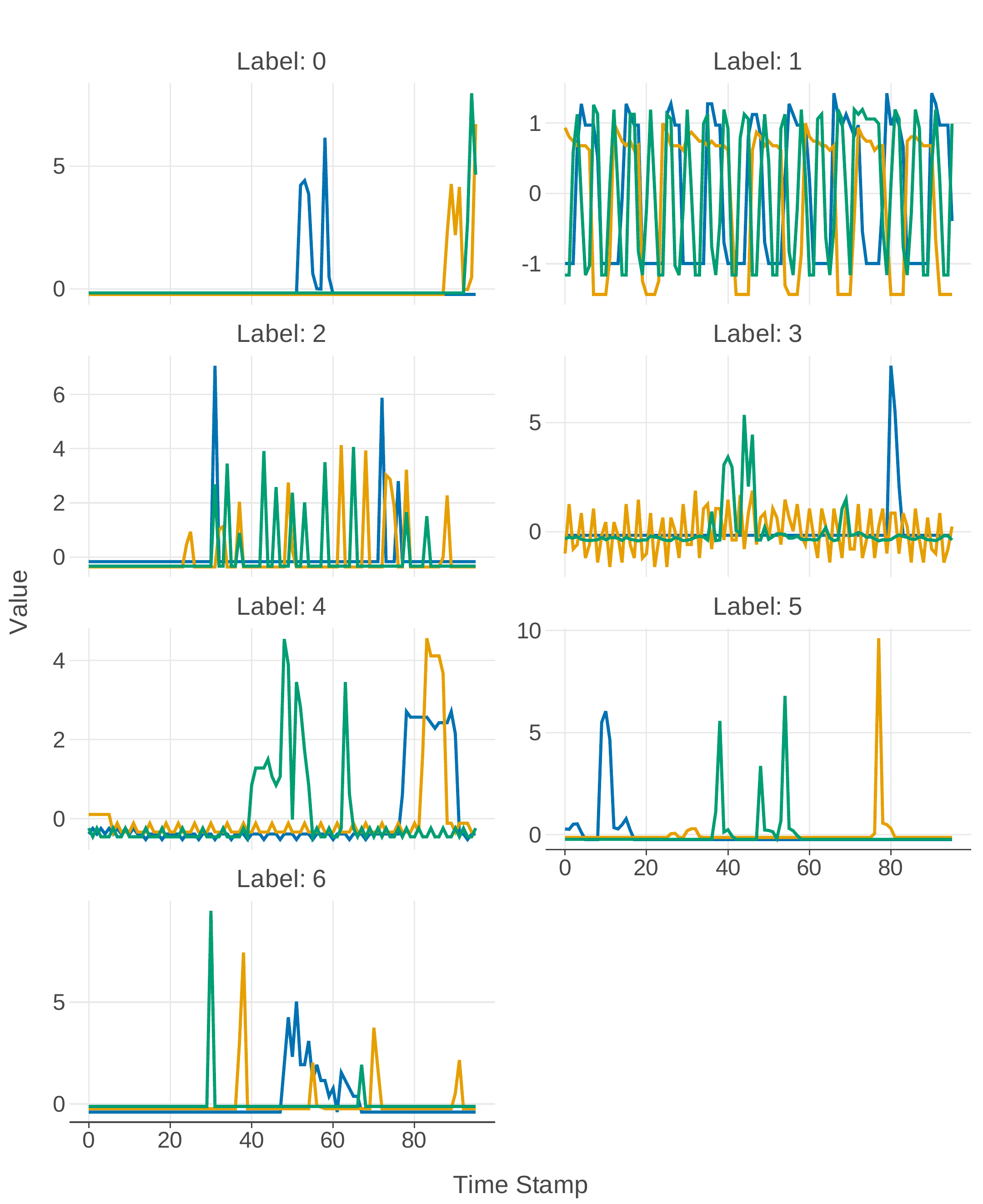}
\caption{Representative samples from the one multiclass classification dataset used in this study, ElecDev. Three randomly selected instances per class are shown to illustrate the characteristic patterns and variability between classes. If datasets had labels not starting from 0, they were remapped to start from 0 (ranging from 0 to $C-1$), preserving the existing order.}\label{fig:samples_multiclass}
\end{figure}
\clearpage

\begin{credits}
\subsubsection{\ackname} This paper is supported by the European Union’s HORIZON Research and Innovation Program under grant agreement No. 101120657, project ENFIELD (European Lighthouse to Manifest Trustworthy and Green AI). 

\subsubsection{\discintname}
All authors declare that they have no conflicts of interest.
\end{credits}

\bibliographystyle{splncs04}
\bibliography{references}

\end{document}